\documentclass[runningheads]{llncs}
\usepackage{graphicx}
\usepackage{breakurl}
\begin{document}
\title{The First Swahili Language Scene Text Detection and Recognition Dataset}
%
%
\author{Fadila Wendigoundi Douamba  \and Jianjun Song \thanks{Co-first author} \and 
Ling Fu \and
Yuliang Liu \thanks{Corresponnding author: ylliu@hust.edu.cn} \and
Xiang Bai }
\authorrunning{F. W. Douamba, J. Song, et al.}
\institute{Huazhong University of Science and Technology
\\
}
\maketitle           
\begin{abstract}
Scene text recognition is essential in many applications, including automated translation, information retrieval, driving assistance, and enhancing accessibility for individuals with visual impairments. Much research has been done to improve the accuracy and performance of scene text detection and recognition models. However, most of this research has been conducted in the most common languages, English and Chinese. There is a significant gap in low-resource languages, especially the Swahili Language. Swahili is widely spoken in East African countries but is still an under-explored language in scene text recognition. No studies have been focused explicitly on Swahili natural scene text detection and recognition, and no dataset for Swahili language scene text detection and recognition is publicly available. We propose a comprehensive dataset of Swahili scene text images and evaluate the dataset on different scene text detection and recognition models. The dataset contains 976 images collected in different places and under various circumstances. Each image has its annotation at the word level. The proposed dataset can also serve as a benchmark dataset specific to the Swahili language for evaluating and comparing different approaches and fostering future research endeavors.
The dataset is available on GitHub via this link: https://github.com/FadilaW/Swahili-STR-Dataset


\end{abstract}
\section{Introduction}
Communicating nowadays heavily relies on textual content. The texts are an excellent means of communication, a means that is also perpetuated over a very long period. Scene text is widespread and contains quite meaningful semantics and information to understand the real world. Diverse services such as newspapers, hospitals, financial services, insurance, and legal institutions increasingly digitise most documents for practical use. Applications such as automotive assistance, industrial automation, robot navigation, real-time scene translation, fraud detection, image retrieval, product search, etc. that use scene text recognition are born and evolve day by day. It is now imperative to be able to understand and interpret the text contained in the images. In addition, text is available everywhere in many critical natural scenes: road signs, advertisements, posters, streets, restaurants, shops, etc.

In recent years, researchers have made significant progress in developing models for detecting and recognizing text in challenging scenarios such as blurred images, unconventional backgrounds, varying lighting conditions, curved text or imagery captured in adverse environmental settings. However, most of the research has been focused on widely spoken languages like English and Chinese, leaving other languages spoken in resource-limited regions such as rural India and Africa with less attention and resources. Consequently, there is a lack of suitable datasets and tailored models for many world languages, making it difficult to effectively address the challenges of detecting and recognizing text in scene images in these languages.

Swahili, also called Kiswahili, is one of the most spoken languages in Africa. Over 100 million people speak Swahili in many African countries, including Tanzania, Uganda, the Democratic Republic of Congo, Burundi and Kenya. The language is the official language in Tanzania and Kenya and is widely spoken in public administration, education, and the media. Swahili has borrowed words from foreign languages like Arabic (around 40 per cent), Persian, Portuguese, English and German.
The Swahili language is still classified as a low-resource language. The natural language processing tasks have been limited due to the scarcity of annotated data. 

Although Swahili uses the Latin script, most large datasets with Latin script focus on languages with different linguistic features like English. This lack of attention leaves Swahili, a language spoken by millions of people, without dedicated resources for optimizing and fine-tuning text detection and recognition models to its unique characteristics. Table 1 presents some of the characteristics of the language compared to the English language.

The main objective of this paper is to develop a comprehensive scene text dataset for the Swahili language: Swahili-text. recognizing the need for specialised datasets, the proposed collection of images is intended to bridge the gap, providing a benchmark for evaluating existing models and helping the research community develop new state-of-the-art methods for Swahili language scene text detection and recognition. 
Swahili-text contains 976 images, most taken from Tanzanian cities and others from social media. These images include shop tags, advertisement banners, posters, and street names. Each image is manually annotated at the word level. To the authors' knowledge, Swahili-Text is the first comprehensive scene text dataset developed for Swahili scene text detection and recognition.

\begin{table}
\caption{Swahili Language characteristics}\label{tab1}
\begin{tabular}{|p{2cm}|p{5cm}|p{5cm}|}
\hline
Aspect & Swahili & English\\
\hline
 Script Complexity & Swahili employs Latin script but includes unique characters(ng) and diacritics, requiring recognition of specific linguistic features. & English has a more straightforward recognition process with familiar characters. \\
 \hline

 Noun Class Prefixes & Swahili's noun class system, indicated by prefixes, adds complexity to recognizing and categorizing nouns correctly. & English lacks a formal noun class system, simplifying the identification of nouns in a scene.\\
 \hline
Ligatures and Word Forms & Swahili exhibits variations in ligatures and word forms, where words may be written as one or split into two, requiring flexible recognition & English generally follows consistent word forms, making segmentation and recognition more straightforward.\\
\hline

 Language Variants and Dialects & Swahili has variations across regions and dialects, introducing potential challenges in recognizing diverse linguistic forms & English has regional variations but is often standardized, making scene text recognition more uniform across dialects.\\
\hline
 - & Swahili words can be found mixed with other English words, and this can constitute an additional challenge in Swahili text recognition & -\\

\hline
\end{tabular}
\\
\end{table}

\section{Related work}

\subsection{Swahili Language Datasets for Natural Language Processing}

The Swahili language is still classified as a low-resource language. The natural language processing tasks have been limited due to the scarcity of annotated data. Nevertheless, with the expansion of deep learning and language models, many datasets are becoming increasingly available for language modelling tasks. One of the most used datasets is the Helsinki dataset ~\cite{ref_article1}, which serves as a linguistic dataset for Swahili language research. It provides a collection of Swahili texts in both not annotated and annotated versions. The dataset is designed to support linguistic analysis, corpus linguistics, and various research endeavours related to the Swahili natural language processing tasks.

Gelas et al. ~\cite{ref_article2} developed an annotated dataset for language modelling tasks. This dataset contains sentences from different Swahili online media platforms and includes sentences from various fields such as sports, general news, family, politics and religion. The total number of unique words is 512,000.\cite{ref_article3} combines the dataset with a Swahili syllabic alphabet and transforms the English word analogy dataset proposed by \cite{ref_article4}. Barack W et al. \cite{ref_article5} developed the Kencorpus Swahili Question Answering Dataset (KenSwQuAD). The dataset addresses the scarcity of question-answering (QA) datasets in low-resource languages, specifically Swahili. The research aims to enhance machine comprehension of natural language in tasks such as internet search and dialogue systems for Swahili speakers. Alexander R et al. \cite{ref_article6} address the lack of speech datasets for low-resource languages like Swahili, specifically for spoken digit recognition. The study develops a Swahili spoken digit dataset and investigates the effect of cross-lingual and multi-lingual pre-training methods on verbal digit recognition.

These datasets aim to facilitate research on Swahili language modelling and natural language processing tasks. However, no comprehensive dataset exists for annotated Swahili scene text images in scene text detection and recognition tasks.

\subsection{Latin Script Scene Text Datasets}
The field of scene text recognition has been influenced by standard datasets, enabling researchers to save a significant amount of time and effort in collecting and annotating data. 
This section outlines popular datasets related to Latin script scene text recognition. ICDAR Datasets are popular in the field of document analysis and recognition. The ICDAR 2013 dataset contains 462 high-resolution natural scene images \cite{ref_article7}, such as outdoor scenes, signs, and posters. It introduces challenges, such as multi-oriented text, varying lighting conditions, and a mix of fonts and text sizes to stimulate the development of robust text recognition algorithms. The ICDAR 2015 Incidental Scene Text dataset comprises 1,670 images captured through Google Glass \cite{ref_article8}. The dataset includes incidental scene text with unconventional text shapes, curved text, and text in different languages. Total-text dataset \cite{ref_article9} is proposed to address the multi-oriented and curved text issue. It contains images with text of varying orientations, primarily curved. The MSRA-TD500 dataset \cite{ref_article10}, which combines English and Chinese words, is also popular. It contains 500 arbitrarily orientated images from real-world scenarios annotated at the sentence level. 
Besides datasets in Latin script, several multilingual datasets have been proposed for multilingual scene text recognition \cite{ref_article11}. However, most of these datasets do not include Swahili.

As for the Swahili scene text recognition, to the best of our knowledge, no public dataset for scene text detection and recognition has been created. Although some datasets for the English language can be used since they share the same script, they are not as effective as a Swahili-specific dataset.

\subsection{Scene Text Detection and Recognition Methods}

The explosion of deep learning techniques has significantly impacted the field of scene text detection and recognition. Deep neural networks have opened up a world of possibilities for scene text detection and recognition, making it possible to extract stronger and discriminating features from text images. 

Text detection and text recognition can be viewed as two independent tasks. During the phase of text detection, the objective is to identify and mark the areas in an input image where text is present. Three main approaches exist: regression-based, part-based, and segmentation-based approaches. Regression-based methods directly regress bounding boxes. By transforming text detection into a regression problem, the model learns to estimate the spatial distribution of text instances, making it well-suited for scenarios where precise localisation of text regions is critical \cite{ref_article12,ref_article13}. Part-based methods identify and associate text parts with word-bounding boxes \cite{ref_article14}. Segmentation-based methods combine pixel-level prediction with post-processing to detect text instances using techniques such as semantic segmentation and MSER-based algorithms \cite{ref_article15,ref_article16,ref_article17}. 

Text recognition involves converting detected text regions into character instances using two main methods: connectionist temporal classification (CTC) models and attention mechanism models.
CTC models use recursive neural networks to compute the conditional probability of labeling sequences based on individual frame predictions \cite{ref_article18}. The process consists of three significant steps: feature extraction from text regions using a convolutional network, label distribution prediction at each frame using a recursive neural network, and post-processing to convert per-frame predictions into the final label sequence. Many other research works are based on this approach\cite{ref_article19}. 

Attention-based methods \cite{ref_article20} have also achieved remarkable results in the field of computer vision, including scene text recognition. Attention mechanisms focus on relevant parts of the input, resulting in more accurate character recognition, particularly in complex or variable contexts. This method uses an encoding structure to extract feature vectors from text regions and a decoder structure to generate character instances.
Xiao et al.  \cite{ref_article21} tackle problems where attention mechanisms produce unrelated information and suggest a way to assess the relevance between attention results and queries. By integrating an Attention on Attention (AoA) mechanism into the text recognition framework, irrelevant attention is eliminated, and the text recognition accuracy is improved.

Despite the remarkable advancements in scene text detection and recognition, the lack of labelled training data is still an obstacle. Deep learning algorithms' capacity to generalise to real-world scenarios is limited by the scarcity of large-scale datasets, including annotated scene text images, especially for low-resource languages or languages that have not been studied.

\section{SWAHILI TEXT DATASET}
\subsection{Dataset Description}
The Swahili Scene Text Detection and Recognition Dataset contains 976 natural scene images that were collected from various sources. The data was collected by researchers specializing in computer vision and natural language processing, as well as Swahili native speakers. The images were collected from diverse locations, including internet sources and direct captures in Tanzanian cities, using a phone camera. This ensured that a representative collection of scenes from Swahili-speaking regions was obtained. 

To ensure the accuracy and relevance of the collected images, rigorous quality control measures were implemented, and special attention was given to eliminating images with uneven lighting and blur. The dataset underwent preprocessing steps to remove images with undesirable quality attributes, and instances with incomplete data were either corrected or excluded from the dataset to maintain data integrity. Each image in the dataset is stored in JPEG format.

Swahili-text dataset contains images depicting natural scenes with Swahili text elements such as street signs, street names, advertisements, shop names, banners, and other signage commonly found in Swahili-speaking regions. To facilitate scene text detection and recognition tasks, the dataset is labeled, and the labelling process was conducted by domain experts to ensure accurate annotation of text regions. Figure 1 shows a selection of images that are included in the dataset. For the recognition tasks, the dataset images have been cropped into 8284 images. Figure 2 shows statistics on Swahili-text cropped images, including the number of images grouped by text length and the distribution of character occurrences.

\begin{figure}
    \centering
    \includegraphics[width=0.6\linewidth]{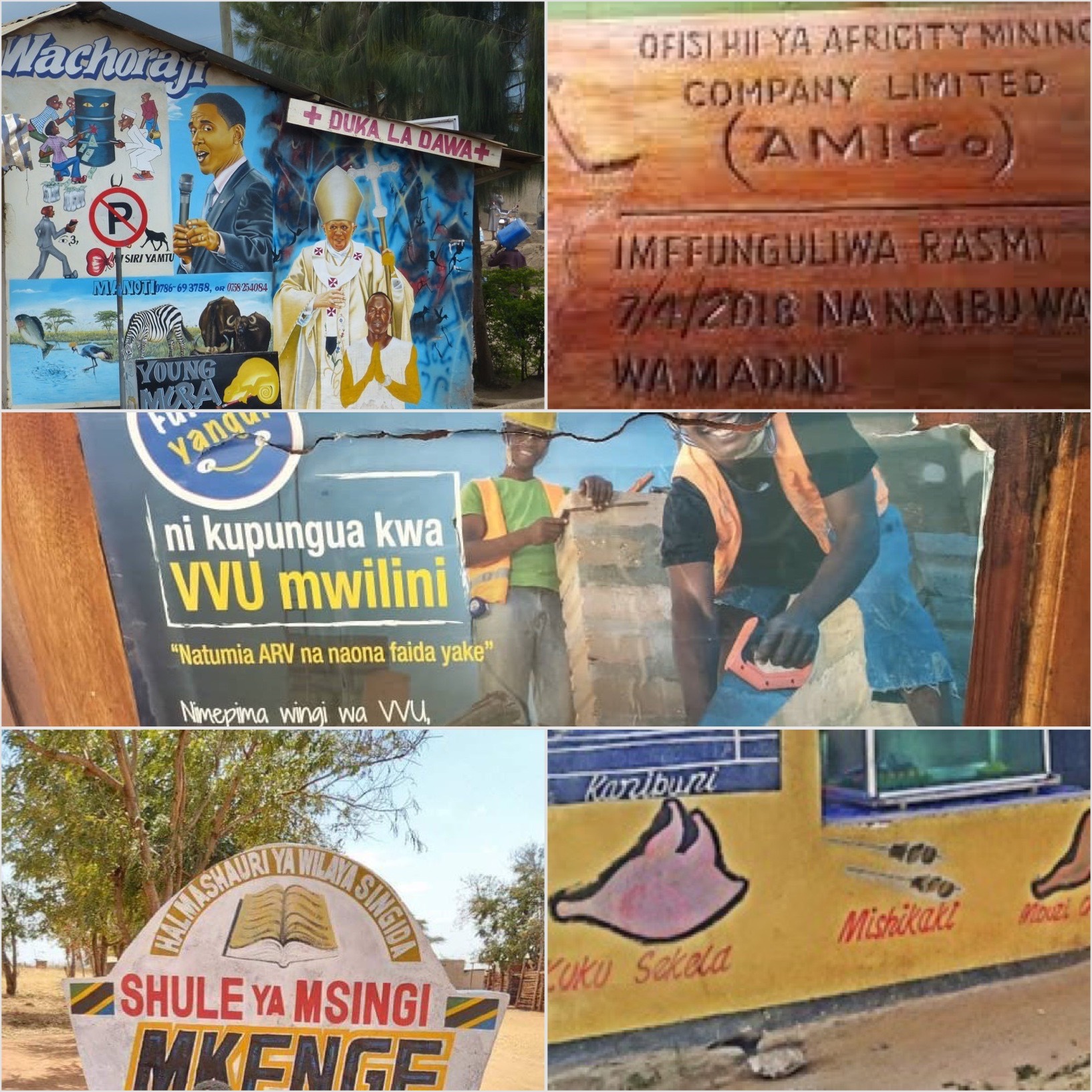}
    \caption{ Dataset Sample}
    \label{fig:enter-label}
\end{figure}

\begin{figure}
    \centering
    \includegraphics[width=0.9\linewidth]{       
   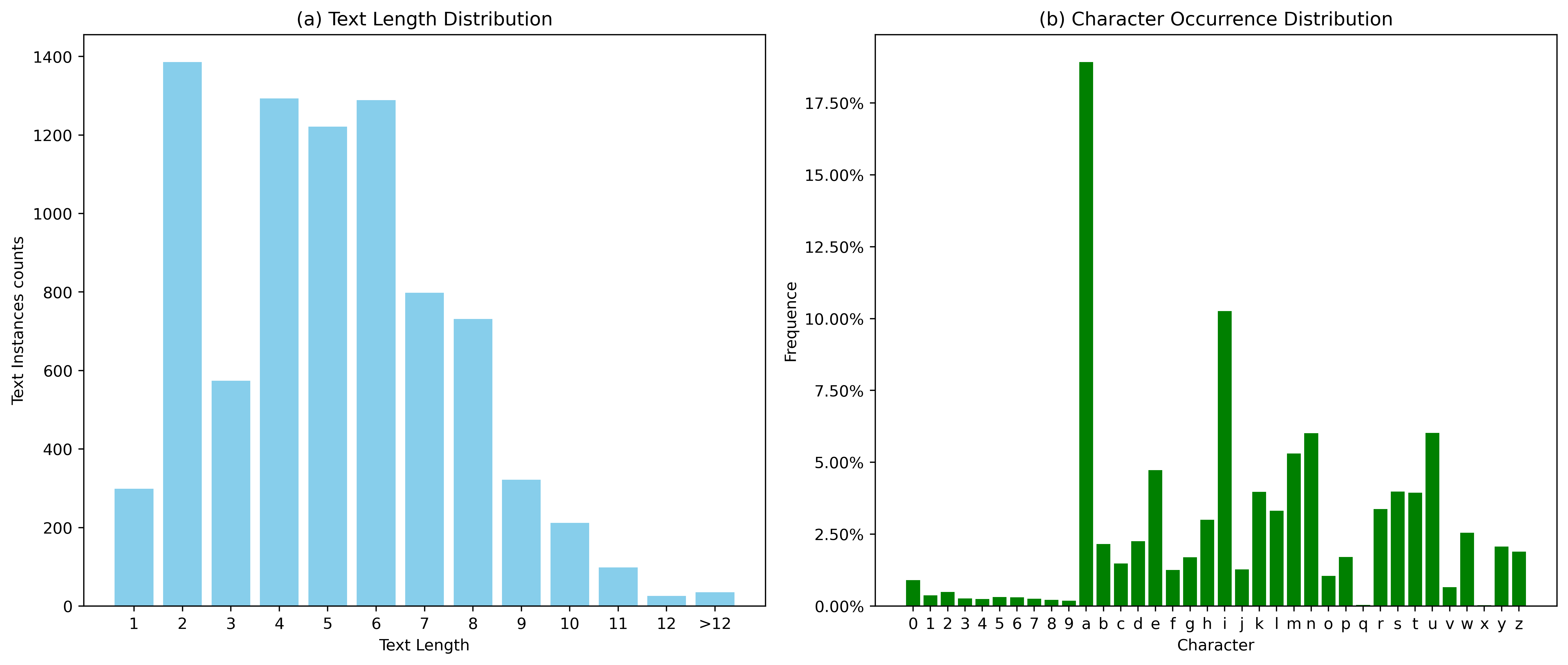}
   \caption{ Statistical analysis of the Swahili text dataset: (a) Grouping images by text length. (b) Counting character occurrences in the dataset.}
    \label{fig:enter-label}
\end{figure}

\subsection{Annotation}

Accurate annotations of text instances are essential for guaranteeing text detection and recognition efficacy and evaluating the system's performance. Consequently, the Swahili-text dataset uses a meticulous manual annotation method. Each text region within every image is annotated with a single bounding box to ensure accuracy, mainly when dealing with the varied shapes and positions of Swahili text.

Each image's text instance annotations are collected into a single file. The bounding box coordinates of a single word and the accompanying text transcription are included in this file. The bounding box is a polygon with n points, each having the coordinates x1 (horizontal position) and y1 (vertical position). Text instances which are not readable are just delimited with the bounding box to contribute to the detection but not the recognition of the text. An example of an annotated image is presented in Figure 3.

\begin{figure}
    \centering
    \includegraphics[width=0.5\linewidth]{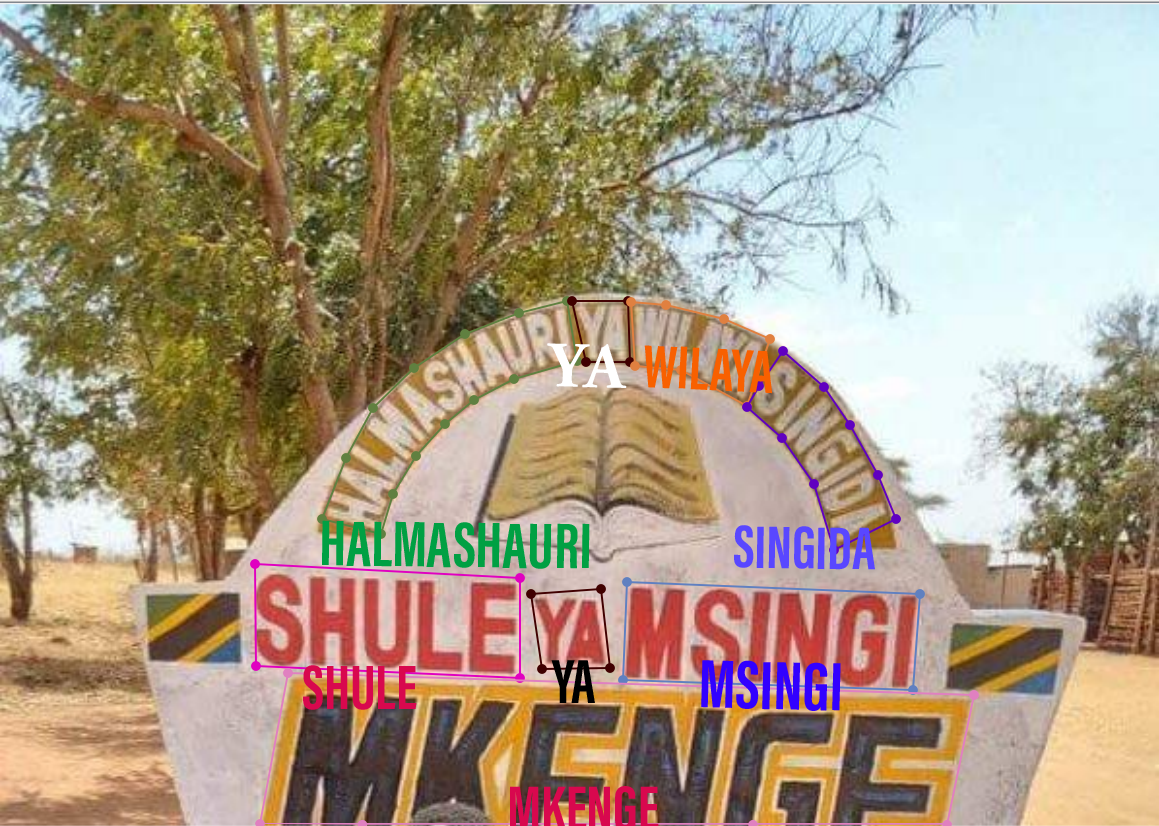}
    \caption{Annotated Image Sample}
    \label{fig:enter-label}
\end{figure}

\section{EXPERIMENTS}
\subsection{IMPLEMENTATION DETAILS}
During our experiments on the Swahili language scene text dataset, we trained six different models for text detection and recognition. For text detection, we used DBNet \cite{ref_article22}, PANet \cite{ref_article23} and FCENet \cite{ref_article24}.For text recognition, we worked with ASTER \cite{ref_article26}, SATRN \cite{ref_article27} and ABINet \cite{ref_article28}. These models were implemented using the PyTorch framework and trained on a single RTX 3090 GPU using different batch sizes for each model. Specifically, we used a batch size of 16 for DBNet, PANet, and FCENet, 1024 for ASTER, 128 for SATRN, and 192 for ABINet.

\textbf{Optimization:} To optimize the performance of the models, various optimization algorithms were used. SGD was used for DBNet and FCENet, Adam for PANet, ABINet, and SATRN, and AdamW for ASTER. The initial learning rates were set to 0.007 for DBNet with a Poly decay strategy, 0.0001 for PANet, 0.003 for FCENet with a decay factor of 0.8 every 200 epochs, 0.0004 for ASTER, $4 \times 10^{-5}$ for SATRN, and $1 \times 10^{-7}$ for ABINet.

\textbf{Training:} The number of training epochs for each model were updated as follows: DBNet (440), PANet (440), FCENet (1120), ASTER (1024), SATRN (366), SATRN (5), ABINet (600).
The pre-trained model used was \texttt{dbnet\_resnet18\_fp\\nc\_1200e\_totaltext-3ed3233c.pth} for DBNet, \texttt{panet\_resnet18\_fpem-ffm\\\_600e\_ctw1500\_20220826\_144818-980f32d0.pth} for PANet, \texttt{fcenet\_resnet50\\\_fpn\_1500e\_totaltext-91bd37af.pth} for FCENet, and \texttt{aster\_resnet45\_6e\_st\\\_mj-cc56eca4.pth} for ASTER. The SATRN pre-trained model used is \texttt{satrn\_sh\\allow\_5e\_st\_mj\_20220915\_152443-5fd04a4c.pth}, and for ABINet, \texttt{abinet\_20e\\\_st\_an\_mj\_20221005\_012617-ead8c139.pth}, all provided by MMOCR \cite{ref_article25}. These pre-trained models are fine-tuned on the training set to obtain the final results.

\subsection{TEXT DETECTION EXPERIMENT}

For the text detection experiment, the dataset has been split into two sets of images: 700 images for training and 276 images for testing. 
In evaluating the effectiveness of our trained models in detecting text within images, we used precision, recall, and F-score as our primary evaluation metrics. Precision measures the accuracy of the positive predictions made by the model, while recall measures the model's ability to correctly identify all relevant instances of text. The F-score, which is the harmonic mean of precision and recall, provides a balanced measure of the model's overall performance.
The results of our experiment on text detection are shown in Table 2, which presents the precision, recall, and F-score metrics for each model. These metrics provide information about the strengths and weaknesses of each model, enabling a comparison of their performance on the dataset. 
In Figure 4, we can see some visualizations that show the text detection capabilities of the FCENet model, which performs the best among the other models.

\begin{table}[htbp]
    \centering
    \caption{Detection experiment results}
    \begin{tabular}{ccccccc}
    \hline
      Method   & Backbone & Venue & \multicolumn{3}{c}{Swahili-text} & \\
      & & & Precision & Recall & F-score & \\
      \hline
         DBNet \cite{ref_article22} & Res18 & AAAI’2020 & 89.8 & 80.0 & 84.6\\
        PANet \cite{ref_article23} &  Res18 & CVPR'2018 & 88.7 & 75.8 & 81.7\\
        FCENet \cite{ref_article24} & Res50 & CVPR 2021 & 90.8 & 79.5 & 84.8\\
        \hline
    \end{tabular}
    \label{tab:my_label}
\end{table}

\begin{figure}
    \centering

    \includegraphics[width=0.5\linewidth]{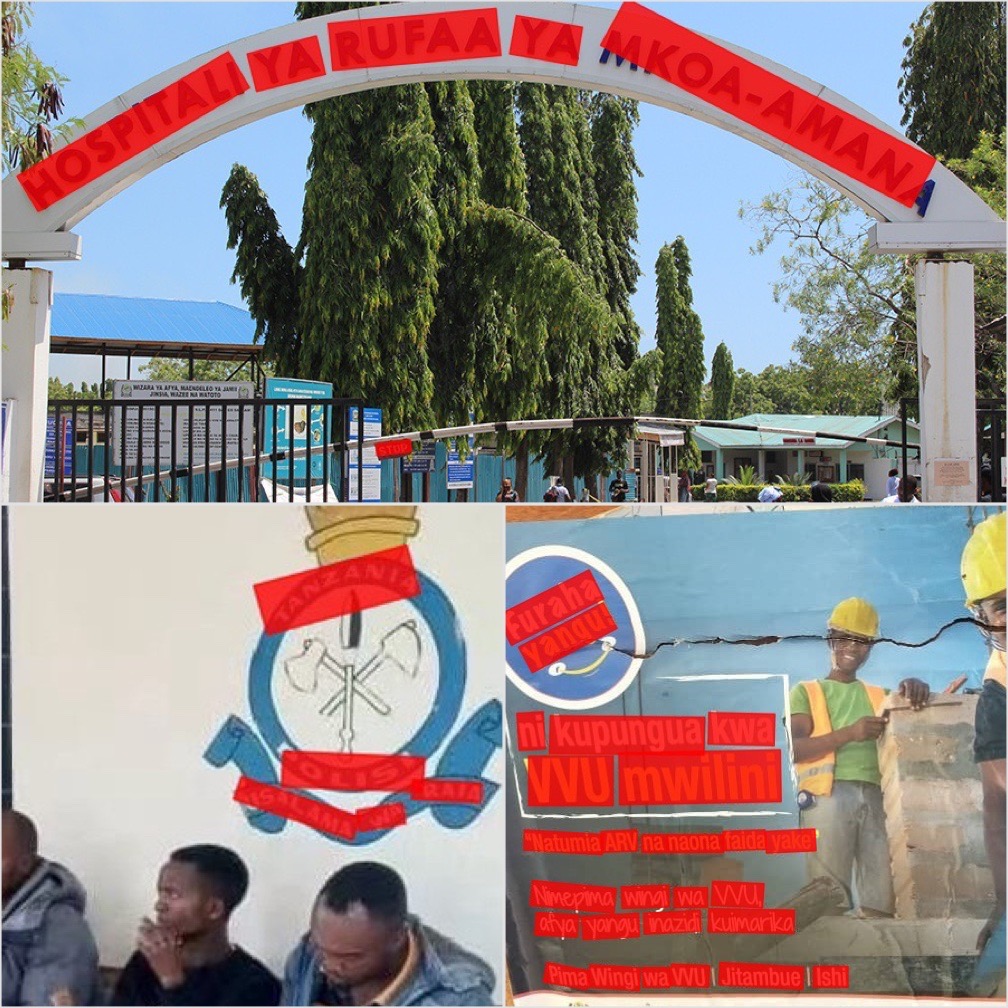}
    \label{fig:enter-label}
        \caption{Text detection visualization}
\end{figure}

\subsection{TEXT RECOGNITION EXPERIMENT}
The dataset for the text recognition experiment was divided into two sets. The training set was composed of 700 images, while the testing set had 276 images. Each image was processed to crop out the text instances, keeping only the higher-quality text portions for analysis. The resulting training set contained 5,870 cropped images, and the test set had 2,414 cropped images that were exclusively text segments. This facilitated a focused analysis of the text recognition task. The evaluation of the text recognition experiment focused on the word-level accuracy, ignoring letter case and symbols. The outcomes of the text recognition experiment are presented in Table 3, which shows the word-level accuracy metrics for each network used. Visualizations were created in Figure 5, which showcases the text recognition capabilities of the SATRN model that achieved the best performance on the dataset. Figure 6 presents some failure cases of the SATRN model in recognizing Swahili words.

\begin{table}
    \centering
       \caption{Text Recognition Experiment Results}
\begin{tabular}{ccc}
\hline
Model & Venue  & Result \\
\hline
ASTER \cite{ref_article26} & TPAMI’2019 & 88.8 \\
SATRN \cite{ref_article27} & CVPRW’2021 & 93.9 \\
ABINet \cite{ref_article28} & CVPR’2021 & 92.7 \\
\hline
\end{tabular}
\\
    \label{tab:my_label}
\end{table}

\begin{figure}
    \centering
    \includegraphics[width=0.75\linewidth]{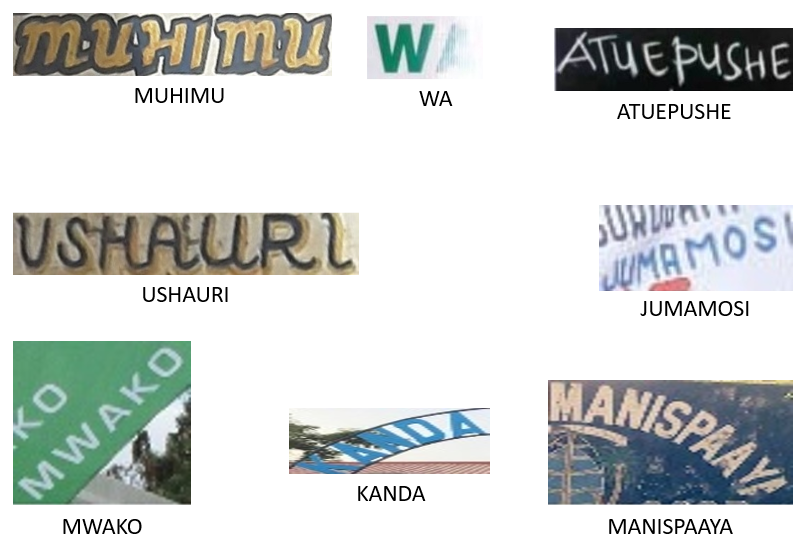}
      \caption{Text Recognition Visualizations}
    \label{fig:enter-label}
\end{figure}

\begin{figure}
    \centering
    \includegraphics[width=0.75\linewidth]{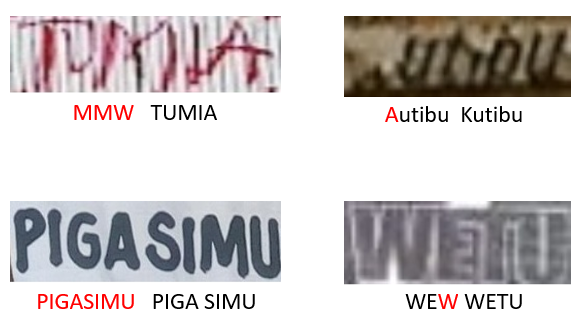}
  \caption{Text Recognition Failure Cases}
    \label{fig:enter-label}
\end{figure}

\section{DISCUSSION}

Based on our findings, the FCENet model had the highest detection F-score of 84.8. Furthermore, the SATRN model outperformed ASTER by more than 0.5, demonstrating its superiority in text recognition accuracy on our dataset. Specifically, it achieved an accuracy of 93.9 on the Swahili-text dataset. We believe that adapting models to the linguistic nuances and characteristics of the target language is crucial for achieving high performance in text detection and recognition tasks. Although the SATRN model had the best performance among the evaluated models, its visualizations showed certain limitations. Some words were not accurately recognized, which indicates the need for more appropriate methods that are specifically designed for Swahili language to achieve optimal performance in Swahili scene text recognition.

\section{CONCLUSION AND FUTURE DIRECTIONS}
We propose a dataset specifically tailored for detecting and recognizing natural scene text in Swahili, an under-explored linguistic domain in current research. The dataset aims to drive advancements in state-of-the-art methods and the development of new techniques to address scene text detection and recognition problems in Swahili language. 
The dataset comprises 976 annotated scene images that can be used for text detection and 8284 cropped images for recognition.
Swahili-text is evaluated using three scene text detection methods: DBNet, PANet, and FCENet.Three text recognition models were used: ABINet, SATRN, and ASTER. After fine-tuning, FCENet had the highest detection accuracy with an F-score of 84.8, while SATRN achieved 93.9 accuracy on the Swahili dataset. The experiments demonstrate that optimal results in Swahili text recognition require adapting the methods to the specific dataset and linguistic characteristics of the language. 
The Swahili Scene Text Recognition dataset stands as a pioneering resource that unlocks new research avenues in Swahili language text detection and recognition within natural scenes, as well as other under-researched language scene text detection and recognition. The dataset not only tackles the difficulties of detecting and reading Swahili text but also sets the foundation for further developments in the field. As the dataset continues to expand and improve, it will undoubtedly push the boundaries of Swahili scene text recognition.

\section*{Acknowledgments}
This work was supported by the National Natural Science Foundation of China (No.62206104, No.62225603).

%
%

%
%
%

\begin{thebibliography}{8}
\bibitem{ref_article1} Hurskainen, A.: Helsinki corpus of Swahili 2.0 (hcs 2.0) annotated version. (2016)

\bibitem{ref_article2} Gelas, H., Besacier, L., Pellegrino, F.: Developments of Swahili resources for an automatic speech recognition system. (2014)

\bibitem{ref_article3} Shikali, C.S., Mokhosi, R.: Enhancing African Low-resource Languages: Swahili Data for Language Modelling. (2020)

\bibitem{ref_article4} Mikolov, T., Karafiát, M., Burget, L., Černocký, J., Khudanpur, S.: Recurrent neural network based language model. Proceedings of the Eleventh Annual Conference of the International Speech Communication Association, (2010)

\bibitem{ref_article5} Wanjawa, B.W., Wanzare, L.D.A., Indede, F., McOnyango, O., Muchemi, L., Ombui, E.: A question answering dataset for Swahili low resource language. (2022)

\bibitem{ref_article6} Kivaisi, A.R., Zhao, Q., Mbelwa, J.T.: Swahili speech dataset development and improved pre-training method for spoken digit recognition. (2023)

\bibitem{ref_article7} Karatzas, D., Shafait, F., Uchida, S., Iwamura, M., Gomez-Bigorda, L., Mestre, S. R., Mas, J., Mota, D. F., Almazan, J. A., De Las Heras, L. P. (2013). ICDAR 2013 robust reading competition. In Proceedings of the International Conference on Document Analysis and Recognition, ICDAR

\bibitem{ref_article8} Karatzas, D., Gomez-Bigorda, L., Nicolaou, A., Ghosh, S., Bagdanov, A., Iwamura, M., Matas, J., Neumann, L., Chandrasekhar, V. R., Lu, S., Shafait, F., Uchida, S., Valveny, E. (2015). ICDAR 2015 Competition on Robust Reading. In 13th International Conference on Document Analysis and Recognition (ICDAR) (pp. 1156-1160)

\bibitem{ref_article9} Chng, C.K., Chan, C.S.: Total-Text: A comprehensive dataset for scene text detection and recognition. Proceedings of the 2017 14th IAPR International Conference on Document Analysis and Recognition (ICDAR)

\bibitem{ref_article10} Yao, C., Bai, X., Liu, W., Ma, Y., TU, Z.: Detecting texts of arbitrary orientations in natural images. 2012 IEEE Conference on Computer Vision and Pattern Recognition (2012)Conference on Computer Vision and Pattern Recognition, 2012.

\bibitem{ref_article11}Nayef, N., Patel, Y., Busta, M., Chowdhury, P. N., Karatzas, D., Khlif, W., Matas, J., Pal, U., Burie, J.-C., Liu, C.-L., Ogier, J.-M. (2019). ICDAR2019 robust reading challenge on multi-lingual scene text detection and recognition—RRC-MLT-2019. In Proceedings of the International Conference on Document Analysis and Recognition (ICDAR) 
  
\bibitem{ref_article12} Liao, M., Shi, B., Bai, X.: TextBoxes: A fast text detector with a single deep neural network. (2016)

\bibitem{ref_article13} Liao, M., Shi, B., Bai, X.: TextBoxes++: A single-shot oriented scene text detector. IEEE Transactions on Image Processing (2018)

\bibitem{ref_article14} Shi, B., Bai, X., Belongie, S., J.: Detecting oriented text in natural images by linking segments, (2017)

\bibitem{ref_article15} Xie, E., Lu, Y., Wang, S., Zuo, X., Luo, P., Wu, X.: Scene text detection with supervised pyramid context network. (2018)

\bibitem{ref_article16} Li, X., Yin, P., Zhang, Z., Jin, L., Wang, C.: Shape robust text detection with progressive scale expansion network. (2018)

\bibitem{ref_article17} Zhang, C., Zhang, Z., Shen, W., Yao, C., Liu, W., Bai, X.: Multi-oriented text detection with fully convolutional networks. Proc. Conf. Comput. Vision Pattern Recognition (2016)

\bibitem{ref_article18} Graves, A., Schmidhuber, J.: Connectionist temporal classification: Labelling unsegmented sequence data with recurrent neural networks. Proceedings of the 23rd International Conference on Machine Learning (2006)

\bibitem{ref_article19} Buoy, R., Iwamura, M., Srun, S., Kise, K.: Scene text recognition using Explainable Connectionist-Temporal-Classification. Journal of Imaging (2023)

\bibitem{ref_article20}Vaswani, A., Shazeer, N., Parmar, N., Uszkoreit, J., Jones, L., Gomez, A.N., Kaiser, Ł., Polosukhin, I.: Attention is All You Need. Proceedings of the 31st Conference on Neural Information Processing Systems (NIPS) (2017)


\bibitem{ref_article21} Xiao, Z., Ma, Y., Jiang, W., Luo, P.: An extended attention mechanism for scene text recognition. Expert Systems With Applications (2022)

\bibitem{ref_article22} Liao, M., Wan, Z., Yao, C., Chen, K., Bai, X.: Real-time Scene Text Detection with Differentiable Binarization. (2019)

\bibitem{ref_article23} Liu, S., Qi, L., Qin, H., Shi, J., Jia, J.: Path Aggregation Network for Instance Segmentation. Proceedings of the 2018 IEEE/CVF Conference on Computer Vision and Pattern Recognition (CVPR), (2018).

\bibitem{ref_article24} Zhu, Y., Wu, L., Zhang, C., Zhou, Y., Wei, Y., Yao, A., Lu, H.: Fourier Contour Embedding for Arbitrary-Shaped Text Detection. in Proceedings of the IEEE Conference on Computer Vision and Pattern Recognition (CVPR), (2021).

\bibitem{ref_article25} Kuang, Z., Liu, W., Han, X., Ding, H., Li, X., Peng, G., Zhang, Y.:Mmocr: A comprehensive toolbox for text detection recognition and understanding, in Proceedings of the 29th ACM International Conference on Multimedia.

\bibitem{ref_article26}Shi, B., Yang, M., Wang, X., Lyu, P., Yao, C., Bai, X.: An attentional scene text recognizer with flexible rectification, IEEE Transactions on Pattern Analysis and Machine Intelligence, (2019)

\bibitem{ref_article27} Lee, J., Park, S., Baek, J., Oh, S. J., Kim, S., Lee, H.: On recognizing texts of arbitrary shapes with 2D self-attention, Conference on Computer Vision and Pattern Recognition Workshops (CVPRW), (2020)

\bibitem{ref_article28} Fang, S., Mao, Z., Xie, H., Wang, Y., Yan, C., Zhang, Y.: ABINet++: Autonomous, bidirectional and iterative language modeling for scene text spotting, IEEE Transactions on Pattern Analysis and Machine Intelligence, (2023)


\end{thebibliography}
%

\end{document}